\documentclass[runningheads]{llncs}
\usepackage[colorlinks,linktoc=all]{hyperref}
\usepackage[all]{hypcap}
\hypersetup{citecolor=MidnightBlue}
\hypersetup{linkcolor=MidnightBlue}
\hypersetup{urlcolor=MidnightBlue}
\usepackage[nameinlink]{cleveref}
\creflabelformat{equation}{#2\textup{#1}#3}  % <- remove parenthesis from equations

\usepackage{amsmath}

\usepackage{amssymb}

\usepackage{color}
\usepackage[usenames,dvipsnames]{xcolor}
\definecolor{shadecolor}{gray}{0.9}

\usepackage{grffile}
\usepackage{graphicx}

\graphicspath{{images/}}

\usepackage{booktabs, array, tabulary}

\newcommand{\colr}[1]{{\color{red} #1}}
\newcommand{\colb}[1]{{\color{blue} #1}}

\begin{document}
\title{Demand forecasting techniques for build-to-order lean manufacturing supply chains}
\author{Rodrigo Rivera-Castro\inst{1} \and
Ivan Nazarov\inst{1} \and
Yuke Xiang \inst{2} \and
Alexander Pletneev \inst{1} \and
Ivan Maksimov \inst{1} \and
Evgeny Burnaev\inst{1}}
\authorrunning{R. Rivera-Castro et al.}
\institute{Skolkovo Institute of Science and Technology, Moscow, Russia
\email{\{rodrigo.riveracastro, alexander.pletneev, ivan.maksimov, e.burnaev\}@skoltech.ru}, \email{\{ivan.nazarov\}@skolkovotech.ru}\\
\and
Huawei Noah's Ark Lab, Hong Kong\\
\email{yuke.xiang@huawei.com}}
\maketitle              % typeset the header of the contribution
\begin{abstract}
Build-to-order (BTO) supply chains have become commonplace in industries such as electronics, automotive and fashion. They enable building products based on individual requirements with a short lead time and minimum inventory and production costs. Due to their nature, they differ significantly from traditional supply chains. However, there have not been studies dedicated to demand forecasting methods for this type of setting. This work makes two contributions. First, it presents a new and unique data set from a manufacturer in the BTO sector. Second, it proposes a novel data transformation technique for demand forecasting of BTO products. Results from thirteen forecasting methods show that the approach compares well to the state-of-the-art while being easy to implement and to explain to decision-makers.

\keywords{Demand Forecasting \and Supply Chain Modelling \and Kernels \and Neural Networks.}
\end{abstract}
\section{Introduction}\label{sec:introduction}
Supply chain management (SCM) represents the managerial backbone of the logistics and production sector. Due to its relevance, new methodologies appear regularly in the literature. One of them is Build-To-Order Supply Chain Management (BTO-SCM). This technique has seen significant adoption. Motivated by the lack of work in the demand forecasting literature on BTO-SCM problems, this research develops and presents \textit{Diagonal Feeding}, a data transformation technique, specially tailored for this setting together with a relevant and novel data set from an electronics manufacturer. Data set and implementations of all methods are available for download \footnote{\url{https://github.com/rodrigorivera/isnn19}}.
%
% --- RESEARCH ABSTRACT ---
%

An accurate demand forecasting is essential for the global economy with stockpiled or in-transit inventories representing 17\% of the world's Gross Domestic Product (GDP), \cite{doi:10.1080/00207543.2018.1524167}. Yet, imprecise demand planning is still pervasive with forecast errors of up to 44-53\% for new products, \cite{Kahn2000AnEI}, \cite{jain2005}. As a result, retailers experience out-of-stock (OOS) events with rates amounting to 8.3\% worldwide, \cite{gruen2002retail}. BTO-SCM helps address the uncertainty that arises from variability in demand.

The aims and backgrounds of this research is to present a technique for data pre-processing accessible to non-technical business experts. Traditional forecasting methods are still being used primarily by over 40\% of demand planners in the industry, \cite{Chase2013}, and the use of novel machine learning methods is a promising area with little academic research and insufficient efforts to expose practitioners to them, \cite{2017arXiv170905548R} \cite{Rivera_2018}.

The significance of developing demand forecasting methods that can be easily adopted by demand planners is evident. For discrepancies as low as 2\%, it is worth improving the accuracy of a forecast, \cite{Fleisch2003}, and a 10\% reduction in OOS can increase retailers' revenue by up to 0.5\%, \cite{wipro2013}. Yet, companies struggle hiring the adequate personnel to address these tasks. For example by 2020, Vietnam is expected to face a shortage of over 500,000 employees with data science and analytic skills and over 80\% of the local workforce is unsuited to fill this gap, \cite{2017apec}. In Europe, over 70\% of surveyed businesses struggle hiring data science personnel and over 60\% are resorting to internal training to upgrade the skills of existing business analysts, \cite{2018esade}. This work seeks to alleviate this situation by presenting a feature engineering technique well-suited for demand forecasting in manufacturing that is both accurate as well as easy to communicate to decision-makers.

The research goal is to propose a method for time series forecast of BTO products that can be adopted by practitioners. For this purpose, the study poses the questions: 1) What is the state of the art in academic research in demand forecasting for BTO products? 2) How does \textit{Diagonal Feeding} support the BTO supply chain? 
To achieve the research goal, two objectives have been assigned: a) To review the existing theory on time series prediction, specially on demand forecasting; b) To make a performance comparison of the proposed technique.
The object of research is the balance between accessibility and precision of methods for time series in an industry setting. The subject of the research is forecasting product demand for BTO-SCM.
%
% --- LITERATURE REVIEW ---
\section{Literature Review} \label{sec:literature_review}
BTO-SCM can be defined according to \cite{Gunasekaran2005} as "the system that produces goods and services based on individual customer requirements in a timely and cost competitive manner by leveraging global outsourcing, the application of information technology and through the standardization of components and delayed product differentiation strategies".
Proponents of BTO supply chains such as \cite{SharmaLaplaca2005} argue that they promote sales, reduce costs and increase customer satisfaction. For example, \cite{Christensen2004} mentions that 74\% of US car buyers would purchase a customized vehicle if the delivery time is less than three weeks. As an example of costs reduction, it is claimed that Nissan, a car manufacturer, could save up to 3600 USD per vehicle, if they were to transition their supply chain completely to BTO. In the technology industry, Dell, a computer maker, has generated up to a 160\% return on its invested capital by implementing a BTO supply chain for its e-commerce website, \cite{Swaminathan2003}.

Demand forecasting has commanded attention from different communities due to its importance in the supply chain management, \cite{Attar2016}. A comprehensive treatment can be found in the works of \cite{Chase2013} and \cite{Gilliand2015}. 
Formally, it can be stated as predicting future values $x_{t+h}$, given a time series $x_{t-w+1},\ldots, x_{t-1}, x_t$, where $w$ is the length of the window on the time series and $h$ is called the prediction horizon, \cite{Manisha2013}. To obtain these predictions, quantitative methods such as ARIMA, exponential smoothing models and alike are often used. Yet, \cite{Ahmed2010} argues that there have been few large scale comparison studies of machine learning models for regression or time series aimed at forecasting problems. 
For the electronics manufacturing industry, \cite{Wan2016} introduced SVM regression to the supply chain of various producers. They concluded that it yields good results compared to other forecasting methods. Although SVM regression is a popular method for forecasting, not everyone has identified it as the most effective method. For example, \cite{Lu2012} used multivariate adaptive regression splines (MARS) to construct sales forecasting models for computer wholesalers. Similarly, \cite{Yelland2010} proposed a Bayesian model for demand forecasting of computer parts and compared it against exponential smoothing and a judgment-based method.
%
% --- DATASET ---
%
\section{Dataset}\label{sec:dataset}

\begin{figure}[!htb]
\minipage{0.5\columnwidth}
  \includegraphics[width=\columnwidth]{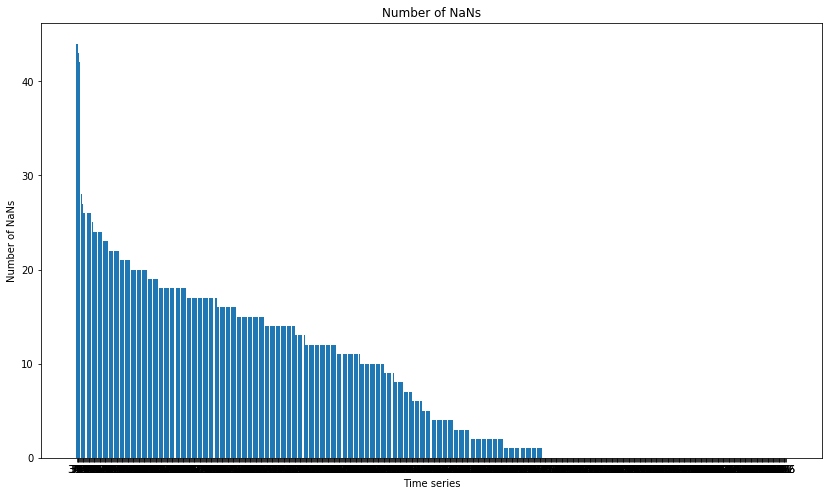}
\endminipage\hfill
\minipage{0.5\columnwidth}
  \includegraphics[width=\columnwidth]{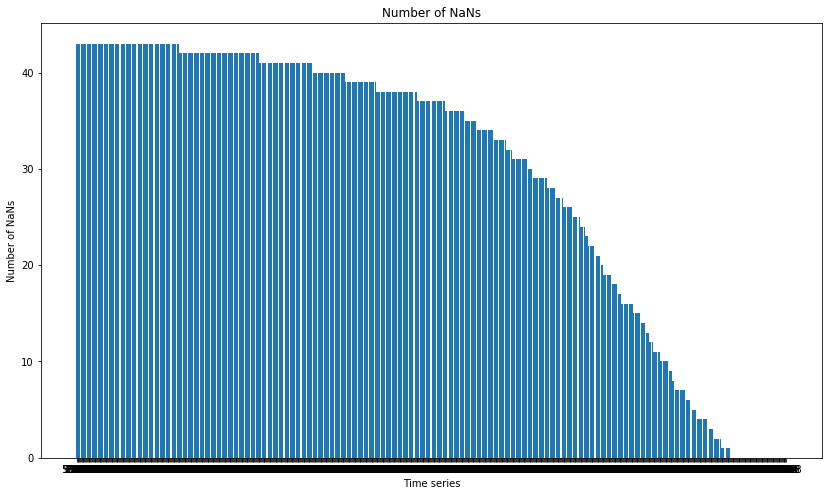}
\endminipage\hfill
\caption{Number of NaNs (zero orders) per item for each of the three data sets. First and Second data set (left), third data set (right). X-axis: Item's ID, Y-axis: Number of zeroes}
\label{fig:tda:orders per item}
\end{figure}

\begin{figure}[!htb]
\minipage{0.5\columnwidth}
  \includegraphics[width=\columnwidth]{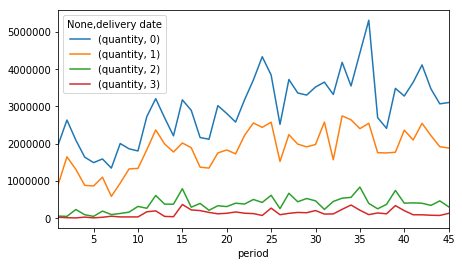}
\endminipage\hfill
\minipage{0.5\columnwidth}
  \includegraphics[width=\columnwidth]{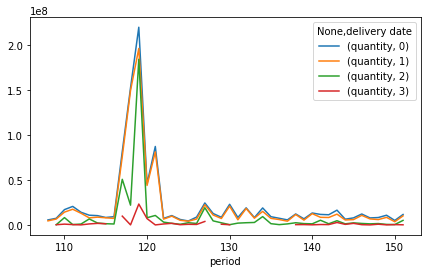}
\endminipage\hfill
\caption{Aggregated quantity by delivery date. First and second data set (left), third data set (right). X-axis: Period, Y-axis: Quantity. Blue line: Same month delivery.}
\label{fig:data:quantity_rpd}
\end{figure}

The data consists of three data sets of observations from an electronics manufacturer representing a subset of their total inventory. Each of them contains the demand of 854 different items, totaling 2562 items, with up to 45 periods of data. The items can be requested for delivery either for the same month or up to three months later. In addition, two hierarchical dependencies are provided in the form of categories and parent-child relationships. These items have varying amounts of required quantities, with many of them being requested only sporadically, as seen in \autoref{fig:tda:orders per item}, and few of them being requested in large quantities.
Noticeable, a significant percentage of products lacks a continuous demand. There are no periods in all three data sets where all 854 products are requested. The demand for items both in variety as well as quantity is significantly higher for the delivery date 0 (same month). From \autoref{fig:data:quantity_rpd}, it can be appreciated that quantities for all periods are consistently higher on delivery date 0 versus others. In the case of delivery date 3, the requested quantities for all periods are significantly smaller. Similarly, the amounts required in the third data set are much higher and they peak out, whereas the first and second data sets have an increasing demand pattern.

%
% --- Diagonal Feeding ---
%
\section{Diagonal Feeding}
\label{sec:diagonal_feeding}
This work introduces the practitioner to \textit{Diagonal Feeding}, a data transformation technique well-suited for multi-step structured forecasting from anticipatory data. As its main benefit, it levers the anticipatory nature of pre-orders' time series data and makes forecasting the pre-order structure more streamlined. As a result, the accuracy of a regression model is improved. 
This is made possible due to the data set containing information not only about the current demand, but also on the volumes of pre-orders made in advance. Advance pre-orders are expectation-driven, naturally forward-looking and known beforehand, as they reflect planning and some anticipation of the market at the end of the period, when the order is to be fulfilled. This information is leveraged for predicting the pre-order structure in
the future by also taking into account the cross-correlations between the pre-orders.

\paragraph{Representation} % (fold)
\label{sub:representation}

Let $q_t^h$ be the total quantity of some item {\it requested by the end of period $t$} through $h$ period advance pre-orders. Those made from the end of $t-h-1$ until the
end of $t-h$. The key property of the data set is that for every item the value $q_t^h$ is {effectively known and available for use} at period $t - h$, i.e. well before $t$. For example, $q_{t-1}^1$ is known at the end of $t-2$ and corresponds to the quantity requested {at the end of $t-1$} accumulated via pre-orders {made by the end of $t-2$}.
In the definition of the data set structure the index $t$ corresponds to the ``{period}'' field, whereas $h$ is the ``{delivery date}'' and $q_t^h$ is the value in the
``{quantity}'' field. The ``{item\_code}'' key is intentionally omitted in order to simplify the explanation of the key aspects of the \textit{Diagonal Feeding} representation.
Prior discussion implies that the $h$-period advance pre-order to be fulfilled at the end of period $t$ seems to reflect certain beliefs about the market environment in period $t$. Therefore, the time series $q_t^{h+1}$ of the volume of period-$t$ pre-orders made $h+1$ periods in advance can be considered to be {\it anticipatory} in relation to the time series $q_t^h$. For example, $q_{t+1}^2$ is known with certainty at $t-1$, represents the total demanded quantity {by the end of $t+1$} accumulated between the end of $t-2$ and the end of $t-1$, and effectively reflects information $2$ periods ahead forward-looking on $q_{t+1}^0$.
In order to be able to utilize the anticipatory nature of time series with different ``delivery dates'' and at the same time to be able to forecast the pre-order structure, the data set in the \textit{Diagonal Feeding} ($t$-frontier) representation is represented as following:
\begin{equation} \label{eq:diag_feed}
  \begin{pmatrix}
      \colb{q_{t+0}^0} & \colb{q_{t+0}^1} & \colb{q_{t+0}^2} & \colb{q_{t+0}^3} \\
      \colr{q_{t+1}^0} & \colb{q_{t+1}^1} & \colb{q_{t+1}^2} & \colb{q_{t+1}^3} \\
      \colr{q_{t+2}^0} & \colr{q_{t+2}^1} & \colb{q_{t+2}^2} & \colb{q_{t+2}^3} \\
      \colr{q_{t+3}^0} & \colr{q_{t+3}^1} & \colr{q_{t+3}^2} & \colb{q_{t+3}^3} \\
      \colr{q_{t+4}^0} & \colr{q_{t+4}^1} & \colr{q_{t+4}^2} & \colr{q_{t+4}^3} \\
  \end{pmatrix}
  \rightarrow
  \begin{pmatrix}
      \colb{x_{t 0}} & \colb{x_{t 1}} & \colb{x_{t 2}} & \colb{x_{t 3}} \\
      \colr{y_{t 0}} & \colb{x_{t 4}} & \colb{x_{t 5}} & \colb{x_{t 6}} \\
      \colr{y_{t 1}} & \colr{y_{t 2}} & \colb{x_{t 7}} & \colb{x_{t 8}} \\
      \colr{y_{t 3}} & \colr{y_{t 4}} & \colr{y_{t 5}} & \colb{x_{t 9}} \\
      \colr{y_{t 6}} & \colr{y_{t 7}} & \colr{y_{t 8}} & \colr{y_{t 9}} \\
  \end{pmatrix}
  \rightarrow
  \begin{pmatrix}
      \colb{z_{0 0}} & \colb{z_{0 1}} & \colb{z_{0 2}} & \colb{z_{0 3}} \\
      \colr{z_{1 0}} & \colb{z_{1 1}} & \colb{z_{1 2}} & \colb{z_{1 3}} \\
      \colr{z_{2 0}} & \colr{z_{2 1}} & \colb{z_{2 2}} & \colb{z_{2 3}} \\
      \colr{z_{3 0}} & \colr{z_{3 1}} & \colr{z_{3 2}} & \colb{z_{3 3}} \\
      \colr{z_{4 0}} & \colr{z_{4 1}} & \colr{z_{4 2}} & \colr{z_{4 3}} \\
  \end{pmatrix}
\end{equation}

\bigskip\par\noindent
In \autoref{eq:diag_feed}, the target $\colr{y_t}$ is the output. Its final representation is the lower diagonal $\colr{z}$. Thus, $\colr{y_t}$ represents the pre-order structure for the next $3$ periods beginning with $t+1$. The objective is to predict the lower diagonal of the matrix by triangularly reshaping the multivariate time series for each ``item'' and introducing some redundancy that localizes relevant anticipatory features. Since the quantity $q_t^h$ is known at time $t-h$, each {diagonal} $(q_{t+s+h}^h)_{h\geq 0}$ in the scheme above is {\it known} at $t+s$, $s\in \mathbb{Z}$.
This representation allows the value $q_s^f$ (the volume of period-$s$ pre-orders made $f$ periods in advance) to be used for forecasting the value $q_t^h$, whenever
$t-h > s-f$, i.e. the moment $s-f$, when $q_s^f$ is revealed, is strictly earlier than the moment $t-h$, when $q_t^h$ becomes known. Therefore, the aim is to predict $\colr{y_t} = \mathrm{cat}\bigl(\colr{q_{t+h}^{:h}} \bigr)_{h\geq 0}$, the pre-order structure for the { next} $4$ periods from $t+1$, based on the preorders $\colb{x_t} = \mathrm{cat}\bigl(\colb{q_{t+h}^{h:}} \bigr)_{h\geq 0}$ and their history, known by $t$, where {\it cat} stands for concatenation of vectors and $q_s^{:f} = (q_s^j)_{j=0}^{f-1}$ ; further, $q_t^{:0}$ is empty. The values in the target $\colr{y_t}$ that seem to be most relevant for practical demand forecasting are on its main diagonal $\colr{z_{pj}}$ with $p = j + 1$, since they are the earliest future volumes.

\paragraph{Correlation analysis} % (fold)
\label{sec:correlation_analysis}
The demand volume data set contains time series for each items with many of them
having sparse volume records due {\it absent periods}. They 
 represent zero-volume orders. Thus, the multivariate time series for each item and ``delivery date'' were rebuilt from the data set by
explicitly setting the volume to zero at absent periods.
To test the viability of the proposed representation, only items with at least $60\%$ periods with actual non-zero recorded volume were kept. After filling the missing periods on the time series of each item $i$ respectively with zeroes, they were transformed into the diagonal representation $(\colb{x_t}^i, \colr{y_t}^i)_{t\in \textit{period}_i}$ and then pooled into one sample $(\colb{x_s}, \colr{y_s})$.
The pre-processed data set was split into development and holdout set, with the latter containing the last $8$ periods available for each item ($134-148$ depending on the particular item).

\paragraph{Results} % (fold)
\label{sub:results}
The Spearman Rank correlation \footnote{\url{http://bit.ly/2zLvNNf}} was computed on the development sub-sample to estimate the ``informativeness''; it is invariant under monotonic transformations of the data and thus insensitive to scale. It was applied to the total demand data set $q_t^h$ for the selected subset of the items. A table containing the triangular grouping of the values is obtained, which the \textit{Diagonal feeding} is built upon. Each value $(p r,\,i j)$ (left-right, top-bottom) in the table is the rank correlation between $q_{t + p}^i$
and $q_{t + r}^j$ computed across {all periods $t$ and items}. For instance, it shows that the most significant correlation ($\geq 0.9$) is between $\colb{x_{11}} = q_{t+1}^1$ with $\colr{y_{10}} = q_{t+1}^0$ computed over pooled sample of different {\it items} and {\it periods}.
Since the values along the ``delivery date'' axis of the data set $q_t^h$ are in fact cumulative (gross), the correlation analysis is applied to the first differences $\delta_t^h = q_t^h - q_t^{h+1}$. Making this difference allows for measuring the pure anticipatory information content of $\colb{x_t}$ in the target $\colr{y_t}$ in the \textit{Diagonal Feeding} representation of $\delta_t^h$.
The correlations of this \textit{differenced} data (net monthly orders) confirms that much of the observed correlation was due to the accumulated (integrated) nature of the $q_t^h$ data. The dramatic drop in the correlation between $\colb{x_{11}}$ and $\colr{z_{10}}$ shows that the orders $\delta_t^0$ of the {\it net current} periods are poorly predictable by the $1$-period ahead net pre-orders $\delta_t^1$ meaning that they are likely to have different drivers. Nevertheless, the relatively small change in values within the sample Spearman rank correlation on the $q_t^h$ data and its counterpart on the $\delta_t^h$ data demonstrates that the $1$ period ahead net pre-orders $\delta_t^1$ and lagged net current period orders $\delta_{t-1}^0$ have generally a little more predictive content for the future pre-order structure than the pre-orders made more than $1$ period in advance.

\subsection{Experimental results} % (fold)
\label{sec:experimental_results}

The \textit{Diagonal Feeding} transforms the demand forecasting problem from a time series analysis setting, where the goal is to model $q^h_t \sim g_h \bigl((q^f_s)_{s-f < t-h}\bigr)$ simultaneously for every $h=0,\,\ldots,\,3$, to a supervised multi-output regression problem, where it seeks to learn $\colr{y_t} \sim g\bigl(\colb{x_t},\, \colb{x_{t-1}},\,\ldots\bigr)$.
Extensive numerical experiments were conducted using various common regression models involving cross-validated grid search over the hyper-parameters. Besides the described pre-process, no further data-engineering was carried out.
The best estimator in each model class was picked via $10$-fold cross-validation on the development set (the first $38$ observed periods of each selected item).
The final test scores were computed on the held out data set using by averaging SMAPE across periods and items:
\begin{equation*}
  \mathrm{s}_j
    = \frac1{\lvert \mathit{items} \rvert}
      \sum_{i \in \mathit{items}}
        \mathrm{s}_{ji}
    \,,\,\quad
  \mathrm{s}_{ji}
    = \frac1{\lvert \mathit{test}_i \rvert}
      \sum_{t\in {\mathit{test}_i}}
        \mathrm{s}_{jit}
    \,,\,\quad
  \mathrm{s}_{jit}
    = \frac{2 \lvert \colr{y^i_{tj}} - \hat{y}^i_{tj} \rvert}
        {\lvert \colr{y^i_{tj}} \rvert + \lvert \hat{y}^i_{tj} \rvert}
    \,,
\end{equation*}
where $\hat{y}^i_{tj}$ is the forecast integer volume of the $j$-th element in the
advance preorder structure of item $i$ in the forecast starting with the future
period $t$.
An assessment of the SMAPE scores of the best models across all the targets in the diagonal representation $z_{pj}$ for $p>j$ reveals that the best model is a collection of $10$ independent large ensembles of the gradient boosted regression trees with $500$ estimators in each ensemble. The next best model with generally lower scores is the $2$-layer dense ReLu network with $(80, 20)$ hidden units each. A histograms of the computed $\mathrm{s}_{ji}$ for each $j$ in the structure of the gradient boosted ensemble will depict that most accurate predictions are for the {next periods's} gross order volume ($q_{t+1}^0$). However, the best prediction accuracy drops dramatically when the full grid search experiment repeated for the \textit{differenced} $q_t^h$ data ($\delta_t^h$). An evaluation of the holdout SMAPE scores for the best models on the $\delta_t^h$ reveals that even the best model, $k$-nearest neighbor regression, completely fails to predict the net volume for the next period.
%
% --- EXPERIMENTS ---
%
\section{Experiments} \label{sec:experiments}
To validate \textit{Diagonal Feeding}, this study carried out an assessment of various methods for demand forecasting. In total, thirteen different methods were assessed. For conciseness, this work focuses on the third data set seen in \autoref{fig:tda:orders per item} and on its delivery date 0 (same period). The evaluated methods are 1) Adaboost, 2) ARIMAX, 3) ARIMA, 4) Bayesian Structural Time Series (BSTS), 5) Bayesian Structural Time Series with a Bayesian Classifier (BSTS Classifier), 6) Ensemble of Gradient Boosting (Ensemble), 7) Ridge regression (Ridge), 8) Kernel regression (Kernel), 9) Lasso, 10) Neural Network (NN), 11) Poisson regression (Poisson), 12) Random Forest (RF), 13) Support Vector Regression (SVR). 

Each of them had as a target value three different options: a) Quantity (non-transformed), b) Log-transformed quantity, c) Min-Max transformed quantity. Additionally, \textit{Diagonal Feeding}, presented in \autoref{sec:diagonal_feeding}, was evaluated for regression methods. Thus, three settings were considered: a) No \textit{Diagonal Feeding}, b) \textit{Diagonal Feeding} with an item by item training (One by One). In this case, a vector containing the input of a specific item is fed individually to a model, c) \textit{Diagonal Feeding} fitting the model on the full data set (All Items). Here, a matrix with the input from all items is used. In all three cases, an individual vector corresponding to a given item is obtained as an output. For a), extensive feature engineering was conducted and 360 features were generated. The specific features are documented in the code base provided. The training set consisted of 37 periods and the test set of 8. To evaluate the performance of the models, the Symmetric Mean Absolute Percent Error (SMAPE) is used. It is defined as $\text{SMAPE} = \frac{200\%}{n} \sum_{t=1}^n \frac{|F_t-A_t|}{|A_t|+|F_t|}$
with $F_t$ being the forecasted value and $A_t$ the actual value at time $t$ respectively. The results reveal that the median SMAPE for all methods is 0,42 and for methods with \textit{Diagonal Feeding} is 0,43. The best Top 10 models had a median SMAPE of 0,31 and the Top 10 for those using Diagonal Feeding exclusively was 0,37.
%
% -- DISCUSSIONS ---
%
\section{Discussion}\label{sec:discussion}
\paragraph{Diagonal Feeding}\label{sec:discussion:df}
The key insight from the analysis of the data set is that the next period's gross total demand volume $q_{t+1}^0$ is mostly determined by the currently known one-period ahead pre-orders for the period
($q_{t+1}^1$). The correlation analysis and the results of the grid search experiment confirm the observation that the net next period's volume $\delta_{t+1}^0$ is the difference between $q_{t+1}^0$ and $q_{t+1}^1$. Viewed through \textit{Diagonal Feeding}, it is mostly independent of the history of net pre-orders for the period $t+1$ and is
thus less predictable from advance pre-order data, as indicated by a correlation analysis and results from a grid search experiment. The success of forecasting the $q_{t+1}^0$, especially in contrast to the
other next period's pre-order volumes $q_{t+1+j}^j$ for $j\in \{1,2,3\}$, might be attributed to the observed high correlation of the one-period ahead pre-order volume $q_{t+1}^1$.
Further, \textit{Diagonal Feeding} delivers results comparable to those obtained doing extensive feature engineering; in this case, more than 300 features were generated. Along these lines, exploring different transformations of the target value is essential. For some models using log-transform or min-max delivered good results. For example, a Neural Network without a transformed quantity fitted on the full data set had a SMAPE of 1,13, with a log transformation, it was 0,38.

\paragraph{Experiments}
The best model was Adaboost with an SMAPE of 0,17. It was followed by the Ensemble of Random Forests with 0,18. Yet, it is important to remember that these models had extensive feature engineering with over 300 features generated. For \textit{Diagonal Feeding}, the best method was a random forest with log transform and fitted on the full data set. It obtained 0,34. This was significantly better than the average of methods trained on the data set with feature engineering, which obtained a significantly higher SMAPE of 0,42.
%
% --- CONCLUSION ---
%
\section{Conclusion} \label{sec:conclusion}
This work introduced \textit{Diagonal Feeding}, a technique specially useful for forecasting Build-To-Order products. It helps improving accuracy whenever future delivery dates are known. This approach does not require domain knowledge, extensive feature engineering or advanced technical skills. The results from multiple experiments show that there is no go-to technique for time series prediction. In addition, this research made available a highly-relevant and novel data set. A challenge in developing methods for demand forecasting of BTO products is the lack of public data sets.
As a further line of work, it is worth exploring the impact on \textit{Diagonal Feeding} of transforming the target variable, as it was shown, certain transformations perform better than others. From an algorithmic point of view the approach can be strengthened with non-parametric pre-processing techniques to filter out anomalies such as in \cite{ConformalMartingales2017}, \cite{ConformalAD2015}, \cite{kNN2017}, \cite{EnsemblesDetectors2015}, including multichannel anomaly detection, \cite{Multichannel2017}, performing online aggregation of different forecasting models via long-term aggregation strategies, \cite{AggregationLongTerm}, along with approaches to model quasi-periodic data, \cite{QuasiPeriodic}, and extraction of trends in the presence of non-stationary noise with long tails, \cite{Degradation2016}, \cite{FBM2016}.
Given the potential of BTO supply chains, it is expected that an increasing number of researchers will direct their attention to this area and add additional entries to the literature. 

\section*{Acknowledgements}
The research in section \ref{sec:literature_review} was partially supported by the Russian Foundation for Basic Research grant 16-29-09649 ofi m. The research presented in other sections was supported by the Mexican National Council for Science and Technology (CONACYT), 2018-000009-01EXTF-00154.
%
% --- Bibliography ---
\bibliographystyle{splncs04}
\bibliography{bib}

\begin{thebibliography}{10}
\providecommand{\url}[1]{\texttt{#1}}
\providecommand{\urlprefix}{URL }
\providecommand{\doi}[1]{https://doi.org/#1}

\bibitem{2018esade}
{Agell}, N., {Carricano}, M.: {Adopcion e impacto del Big Data y Advanced
  Analytics en España}. ESADE Business and Law School  (May 2018)

\bibitem{Ahmed2010}
Ahmed, N.K., Atiya, A.F., {El Gayar}, N., El-Shishiny, H.: {An empirical
  comparison of machine learning models for time series forecasting}.
  Econometric Reviews  \textbf{29}(5) (2010)

\bibitem{EnsemblesDetectors2015}
Artemov, A., Burnaev, E.: Ensembles of detectors for online detection of
  transient changes. In: Proc. SPIE. vol.~9875 (2015)

\bibitem{Degradation2016}
Artemov, A., Burnaev, E.: Detecting performance degradation of
  software-intensive systems in the presence of trends and long-range
  dependence. In: IEEE 16th ICDMW. pp. 29--36 (2016)

\bibitem{FBM2016}
Artemov, A., Burnaev, E.: Optimal estimation of a signal perturbed by a
  fractional brownian noise. Theory of Probability \& Its Applications  (2016)

\bibitem{QuasiPeriodic}
Artemov, A., Burnaev, E., Lokot, A.: Nonparametric decomposition of
  quasi-periodic time series for change-point detection. In: Proc. SPIE.
  vol.~9875 (2015)

\bibitem{Attar2016}
Attar, K.S.: {Regression for Demand Forecasting}  \textbf{5}(1) (2016)

\bibitem{doi:10.1080/00207543.2018.1524167}
Bogataj, D., Bogataj, M.: Npv approach to material requirements planning theory
  - a 50-year review of these research achievements. International Journal of
  Production Research  (2018)

\bibitem{Multichannel2017}
Burnaev, E.V., Golubev, G.K.: On one problem in multichannel signal detection.
  Problems of Information Transmission  \textbf{53}(4),  368--380 (Oct 2017)

\bibitem{Chase2013}
Chase, C.W.: {Demand-Driven Forecasting}. John Wiley {\&} Sons, Inc., Hoboken,
  NJ, USA (aug 2013)

\bibitem{Christensen2004}
Christensen, W.J., Germain, R.: Build-to-order and just-in-time as predictors
  of applied supply chain knowledge and market performance. Journal of
  Operations Management  \textbf{23}(5) (nov 2004)

\bibitem{Fleisch2003}
Fleisch, E., Tellkamp, C.: {Inventory inaccuracy and supply chain performance:
  a simulation study of a retail supply chain}. International Journal of
  Production Economics  \textbf{95}(3),  373--385 (mar 2005)

\bibitem{Gilliand2015}
Gilliand, M.: {Business Forecasting}. John Wiley {\&} Sons, Inc., Hoboken, NJ,
  USA (dec 2015)

\bibitem{gruen2002retail}
Gruen, T.W., Corsten, D.S., Bharadwaj, S.: Retail out of stocks: A worldwide
  examination of extent, causes, and consumer responses  (2002)

\bibitem{Gunasekaran2005}
Gunasekaran, A., Ngai, E.: Build-to-order supply chain management: A literature
  review and framework for development. Journal of Operations Management
  \textbf{23},  423--451 (07 2005)

\bibitem{kNN2017}
Ishimtsev, V., Bernstein, A., Burnaev, E., Nazarov, I.: Conformal k-nn anomaly
  detector for univariate data streams. Proceedings of Machine Learning
  Research, vol.~60, pp. 213--227. PMLR, Stockholm, Sweden (13--16 Jun 2017)

\bibitem{jain2005}
Jain, C.: Benchmarking new product forecasting. The Journal of Business
  Forecasting  (2005)

\bibitem{Kahn2000AnEI}
Kahn, K.B.: An exploratory investigation of new product forecasting practices
  (2002)

\bibitem{wipro2013}
Kaul, R.: Retail out-of-stock management: An outcome-based approach  (2013)

\bibitem{AggregationLongTerm}
Korotin, A., V'yugin, V., Burnaev, E.: Aggregating strategies for long-term
  forecasting. Proceedings of Machine Learning Research, vol.~91, pp. 63--82.
  PMLR (11--13 Jun 2018)

\bibitem{Lu2012}
Lu, C.J., Lee, T.S.: {Sales forecasting for computer wholesalers: A comparison
  of multivariate adaptive regression splines and artificial neural networks}.
  Decision Support Systems  (2012)

\bibitem{Manisha2013}
Manisha, G., Vijayalakshmi, M.: {Inter Time Series Sales Forecasting}. Jascse
  \textbf{2}(1) (2013)

\bibitem{2017apec}
{Pompa}, C., {Burke}, T.: {Data Science and Analytics Skills Shortage:
  Equipping the APEC Workforce with the Competencies Demanded by Employers}.
  APEC Human Resource Development Working Group  (2017)

\bibitem{2017arXiv170905548R}
{Rivera}, R., {Burnaev}, E.: {Forecasting of commercial sales with large scale
  Gaussian Processes}. Proc. IEEE 17th ICDMW pp. 625--634 (2017)

\bibitem{Rivera_2018}
Rivera, R., Nazarov, I., Burnaev, E.: Towards forecast techniques for business
  analysts of large commercial data sets using matrix factorization methods.
  Journal of Physics: Conference Series  \textbf{1117},  012010 (nov 2018)

\bibitem{ConformalAD2015}
Safin, A., Burnaev, E.: Conformal kernel expected similarity for anomaly
  detection in time-series data. Advances in Systems Science and Applications
  (2017)

\bibitem{SharmaLaplaca2005}
Sharma, A., Laplaca, P.: Marketing in the emerging era of build-to-order
  manufacturing. Industrial Marketing Management  \textbf{34},  476--486 (07
  2005)

\bibitem{Swaminathan2003}
Swaminathan, J.M., Tayur, S.R.: Models for supply chains in e-business.
  Management Science  \textbf{49}(10),  1387--1406 (oct 2003)

\bibitem{ConformalMartingales2017}
Volkhonskiy, D., Burnaev, E., Nouretdinov, I., Gammerman, A., Vovk, V.:
  Inductive conformal martingales for change-point detection. Proceedings of
  Machine Learning Research, vol.~60, pp. 132--153. PMLR (2017)

\bibitem{Wan2016}
Wan, X.l., Zhang, Z.: {Exploring an Interactive Value-Adding Data-Driven Model
  of Consumer Electronics Supply Chain Based on Least Squares Support Vector
  Machine}. Scientific Programming  (2016)

\bibitem{Yelland2010}
Yelland, P.M., Kim, S., Stratulate, R.: {A Bayesian Model for Sales Forecasting
  at Sun Microsystems}. Interfaces  \textbf{40}(2),  118--129 (2010)

\end{thebibliography}
\end{document}